\documentclass{article}
\usepackage{PRIMEarxiv}
\usepackage{float}
\usepackage[utf8]{inputenc} 
\usepackage[T1]{fontenc}    
\usepackage{hyperref}       
\usepackage{url}            
\usepackage{booktabs}       
\usepackage{amsfonts}       
\usepackage{nicefrac}       
\usepackage{microtype}      
\usepackage{lipsum}
\usepackage{fancyhdr}       
\usepackage{graphicx}       
\graphicspath{{media/}}     
\usepackage{soul,color}
\usepackage{multirow}
\usepackage{amsmath}
\usepackage{amsbsy}
\usepackage{amssymb}
\usepackage{caption}
\usepackage{subcaption}
\usepackage{subfloat}

\usepackage{bm}
\title{LMM-Regularized CLIP Embeddings for Image Classification
\thanks{\textit{\underline{Citation}}: 
\textbf{M. Tzelepi, V. Mezaris, "LMM-Regularized CLIP Embeddings for Image Classification", Proc. IEEE International Symposium on Multimedia (ISM), accepted for publication.}}
\thanks{(c) IEEE. This is the authors' accepted version. The final IEEE-published version can be found in IEEE Xplore. }
}

\author{
  Maria Tzelepi and Vasileios Mezaris\\ 
  Information Technologies Institute (ITI) \\
  Centre of Research and Technology Hellas (CERTH) \\
  Thessaloniki, Greece\\
  \texttt{\{mtzelepi,bmezaris\}@iti.gr} \\
}

\begin{document}
\maketitle
\begin{abstract}
In this paper we deal with image classification tasks using the powerful CLIP vision-language model. Our goal is to advance the classification performance using the CLIP's image encoder, by proposing a novel Large Multimodal Model (LMM) based regularization method. The proposed method uses an LMM to extract semantic descriptions for the images of the dataset. Then, it uses the CLIP's text encoder, frozen, in order to obtain the corresponding text embeddings and compute the mean semantic class descriptions. Subsequently, we adapt the CLIP's image encoder by adding a classification head, and we train it along with the image encoder output, apart from the main classification objective, with an additional auxiliary objective. The additional objective forces the embeddings at the image encoder's output to become similar to their corresponding LMM-generated mean semantic class descriptions. In this way, it produces embeddings with enhanced discrimination ability, leading to improved classification performance. The effectiveness of the proposed regularization method is validated through extensive experiments on three image classification datasets. 
\end{abstract}

\keywords{CLIP \and LMM  \and MiniGPT-4 \and regularization \and image classification}

%
%

\section{Introduction}\label{sec:intro}

CLIP \cite{radford2021learning} is among the most popular Vision-Language Models (VLMs), which has provided outstanding performance in a plethora of vision recognition tasks \cite{ao2023gesturediffuclip,zanella2024delving}. It consists of an image and a text encoder, and it is trained with 400 million image-text pairs in order to predict which of the possible pairs are correct. Apart from its significant zero-shot classification performance, CLIP is used in the literature as a feature extractor for propagating the extracted image embeddings to a classifier and evaluating the performance on datasets of interest, or by finetuning the entire model or the image encoder on the datasets of interest. Besides, VLMs such as CLIP, have allowed for employing the powerful Large Language Models (LLMs) \cite{zhao2023survey} for vision recognition tasks, forming the Large Multimodal Models (LMMs) (also known as Multimodal Large Language Models) \cite{yin2023survey}. For instance, MiniGPT-4 \cite{zhu2023minigpt}, utilizing only a single projection layer, aligns a frozen visual encoder with a frozen LLM utilizing a projection layer.

In this paper, we deal with image classification tasks, i.e., tasks of assigning class labels to images based on their visual content, using the powerful CLIP model. More specifically, we aim to improve the performance achieved by CLIP, finetuned for image classification tasks, exploiting knowledge encoded in LMMs. To accomplish this goal, we first prompt an LMM in a multimodal fashion for obtaining semantic descriptions for the images of the dataset. Then, we obtain the corresponding text embeddings by propagating the responses to the CLIP's text encoder, and we compute for each class a mean semantic class description. Subsequently, apart from adding a fully connected layer and training the image encoder of CLIP for a specific classification task with the class labels, using a conventional supervised loss, we propose to add an additional auxiliary objective. This objective introduces the knowledge derived from the LMM to the training process. Specifically, it forces the image embeddings at the output of the image encoder to become similar to their corresponding LMM-generated mean semantic class description. The additional objective acts as a regularizer that enhances the discrimination ability of the image embeddings. 

Generally, multitask learning has been utilized in the literature as a regularization technique for improving the generalization ability of deep learning models \cite{tzelepi2019graph}. In this paper, we combine regularization objectives that target at enhanced discrimination ability with the emerging capabilities of LMMs. It should be noted that the multimodal nature of the CLIP model, which is pretrained with a contrastive loss in order to find a joint embedding over the image-text pairs, allows for introducing the text-based semantic information encoded in the LMM to the training process. In this way, we can indeed achieve improved discrimination ability of the image embeddings of the CLIP model, leading in turn to advanced classification performance, as it is experimentally validated. 

The rest of the manuscript is organized as follows. Section \ref{sec:Prior} summarizes previous relevant works. Section \ref{sec:method} presents in detail the proposed LMM-regularized CLIP embeddings for image classification tasks. Subsequently, Section \ref{sec:exp} provides the experimental evaluation of the proposed method. Finally, the conclusions are drawn in Section \ref{sec:con}.

%
%

\section{Prior Work}\label{sec:Prior}

In this section we provide a brief review of previous works that involve CLIP and LLMs/LMMs for image recognition tasks. First, a method called Context Optimization (CoOp), is proposed
in \cite{zhou2022learning} for identifying the right prompt with respect to the deployment of VLMs, such as CLIP. CoOp proposes to adapt the prompts for downstream image recognition tasks, utilizing recent advances in prompt learning in natural language processing. Subsequently, a method called Conditional Context Optimization (CoCoOp) \cite{zhou2022conditional} aims to improve the generalization ability of the aforementioned learned context.  

Subsequently, a method called Language Guided Bottlenecks (LaBo) is proposed in \cite{yang2023language} in order to develop interpretable-by-design classifiers. Specifically, LaBo combines GPT-3 \cite{brown2020language} and CLIP, proposing to prompt the first one to generate candidate concepts that describe each of the classes of the classification task, selecting subsequently a subset of discriminating concepts that in turn are aligned to images using CLIP. Finally, a linear layer is applied on the similarity scores in order to learn a weight matrix that represents the importance of each concept in the classification task. Next, in \cite{maniparambil2023enhancing} the GPT-4 model \cite{achiam2023gpt} is used in a unimodal fashion for producing text that is visually descriptive for the classes of the dataset. Then, this information is utilized so as to adapt CLIP to downstream tasks, accomplishing significant improvements in CLIP's zero-shot accuracy, compared to the default CLIP prompt. In addition, CLIP-A-self is proposed, that is a self-attention-based few-shot adapter, in order to exploit this additional information. 

Finally, a method that uses knowledge encoded in LMMs for addressing image classification tasks is proposed in \cite{tzelepi2024exploiting}. Specifically, MiniGPT-4 is used in a multimodal fashion in order to extract semantic information about the images of the dataset. Then, the CLIP's text encoder is used in order to obtain the corresponding text embeddings and use them along with the image embeddings extracted from the CLIP's image encoder, concatenated, in order to train a fully connected layer. Subsequently, the aforementioned method is extended in order to address the disturbing image detection task\cite{tzelepi2024disturbing}. Specifically, apart from the generic semantic descriptions, the LMM is also prompted in order to extract elicited emotions for each image of the dataset, and the corresponding embeddings are additionally used to address the aforementioned task. 

It is evident, surveying the relevant literature, that the vast majority of the relevant works either use the unimodal GPT-3 model for textual prompting (e.g., \cite{yang2023language}) or the multimodal GPT-4 model for textual prompting too (e.g., \cite{maniparambil2023enhancing}), apart from the one in \cite{tzelepi2024exploiting}. Furthermore, all the aforementioned methods, apart from the latter one, obtain class-specific information from the LLM/LMM. On the contrary, in this paper we use the LMM in a multimodal fashion, obtaining sample-specific information. Furthermore, as opposed to \cite{tzelepi2024exploiting}, the proposed method exploits LMM-encoded knowledge in a more effective way, by introducing it to the training process and producing embeddings with enhanced discrimination ability, as it also experimentally validated through the performed experiments.

%
%

\section{Proposed Method}\label{sec:method}
\begin{figure*}
\centerline{\includegraphics[width=0.7\textwidth]{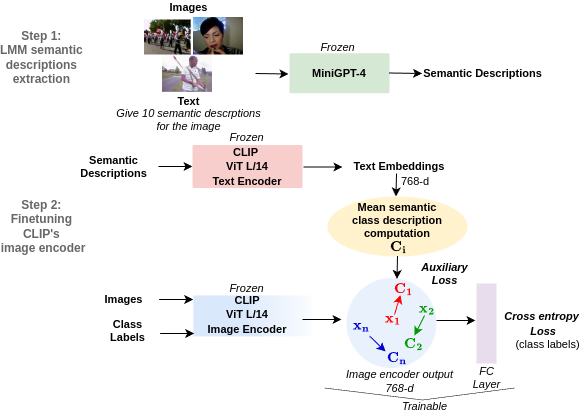}}
\caption{Proposed method: First we use MiniGPT-4 to extract semantic descriptions for each image of the dataset. Then, we use the CLIP's text encoder (frozen) in order to extract the corresponding text embeddings. Subsequently, we compute the mean semantic class descriptions. Finally, we modify the CLIP's image encoder by attaching a fully connected layer at the output of the encoder, and we train the modified model with the class labels using the cross entropy loss, along with an additional auxiliary objective that forces the image embeddings at the penultimate layer of the modified model to become similar to their corresponding mean class description.}
\label{fig:proposed-method}
\end{figure*}

An overview of the proposed method is provided in Fig. \ref{fig:proposed-method}. Specifically, we consider a set of $N$ images $\mathcal{I}=\{\mathbf{I}_i \in \Re^{h\times w \times ch} | i=1, \dots, N\}$, where $h, w, ch$ denote the height, width, and channels of the image, respectively. Each image is associated with a class label $l \in \{0, \dots, N_c - 1$\}, where $N_c$ corresponds to the number of the classes of the considered classification task. We also consider the MiniGPT-4 model, and the ViT-L-14 CLIP version, denoting as $\textit{f}$ and $\textit{g}$ its image and text encoders, respectively. 

In the first stage, we use the MiniGPT-4 model in order to extract semantic descriptions for the images of the dataset. To do so, we prompt the model in a multimodal fashion, using each image along with the text  \textquotedblleft \textit{Give 10 semantic descriptions for the image}\textquotedblright. We denote as $\{s_i^j, i=1, \dots, N ; j=1,\dots,10\}$ the LMM-generated semantic descriptions.

In the second stage, we first use the CLIP's text encoder $\textit{g}$, frozen, in order to obtain the text embeddings of the obtained MiniGPT-4 responses. That is, $\mathbf{t}_i^j=\textit{g}(s_i^j)  \in \Re^{D}$, where $D$ corresponds to the dimension of the embeddings (D=768 in our experiments, since we have utilized ViT-L-14 CLIP version). Then, we use the obtained text embeddings in order to compute the mean text embeddings for each of the classes of the considered problem, i.e., the mean semantic class descriptions. That is: 
\begin{equation}
    \mathbf{C}_l =\frac{1}{|\mathcal{K}_l|}\sum_{\substack{i,j \\ \mathbf{t}_i^j\in \mathcal{K}_l}}\mathbf{t}_i^j,
\end{equation}
where $\mathcal{K}_l$ denotes the set of text embeddings belonging to the class $l$. Subsequently, we consider the CLIP's image encoder $\textit{f}$ and the image embeddings at the output of the encoder for each image of the dataset, i.e., $\mathbf{x}_i=\textit{f}(\mathbf{I}_i)  \in \Re^{D}$. In order to address the considered image classification task, we adapt the CLIP's image encoder by attaching a fully connected layer with $N_c$ neurons at its output. Subsequently, instead of simply finetuning the modified CLIP's image encoder with the class labels using the cross entropy loss, we propose to also train it with an additional auxiliary objective. The proposed objective forces the image embeddings at the penultimate layer of the modified encoder (i.e., the output of the original image encoder of CLIP) to become similar to their corresponding LMM-generated mean semantic class description. That is, the additional objective, $\mathcal{J}_{reg}$, is to minimize the Euclidean distance between the image embedding $\mathbf{x}_i$ at the output of the image encoder and its corresponding LMM-generated mean class description $\mathbf{C}_l^i$, for a sample $i$. This is formulated as follows: 
\begin{equation}\label{eq:reg}
    \min_{\mathbf{x}_i} \mathcal{J}_{reg}=\min_{\mathbf{x}_i}  \sum_{i=1}^{N} \lVert\mathbf{x}_i-\mathbf{C}_l^i \rVert^2_2.
\end{equation}
Thus, the modified image encoder is trained to minimize the overall loss, $\mathcal{J}_{total}$, of the main cross entropy loss, $\mathcal{J}_{ce}$, and the auxiliary regularization loss, $\mathcal{J}_{reg}$. That is: 
\begin{equation}\label{parama}
\mathcal{J}_{total} =  \mathcal{J}_{ce} + \alpha  \mathcal{J}_{reg},
\end{equation}
where the parameter $\alpha$ controls the relative importance of the contributed losses. Note that we train only the output of image encoder and the fully connected layer for computational reasons. Thus, optimizing eq. (3) apart from the classes' separability due to the main classification objective, leads to more compact classes based on the knowledge derived from the LMM. 

%
%

\section{Experiments}\label{sec:exp}

\subsection{Datasets and Evaluation Metrics}
We use three datasets to evaluate the effectiveness of the proposed method: UCF-101 \cite{soomro2012ucf101}, Event Recognition in Aerial videos (ERA) \cite{eradataset}, and Biased Action Recognition (BAR) \cite{nam2020learning}. The UCF-101 dataset, according to \cite{yang2023language}, consists of a training set of 9,537 images and a test set of 3,783 images, divided into 101 action categories. The ERA dataset contains 1,473 training images and 1,391 test images, divided into 25 classes. Finally, the BAR dataset contains 1,941 training images and 654 test images, divided into 6 classes. We use test accuracy in order to validate the effectiveness of the proposed regularization method.

\subsection{Implementation Details}
All the models are trained for 100 epochs using a mini-batch size of 64 samples, the learning rate is set to $1e-4$, while the parameter $\alpha$ in eq. (\ref{parama}) for controlling the relative importance of the two losses is set to $1e-2.$ All the experiments were conducted on an NVIDIA GeForce RTX 3090 with 24 GB of GPU memory.

\subsection{Experimental Results}
\begin{figure*}
\begin{subfigure}{0.3\textwidth}
\centering
\includegraphics[width=\textwidth]{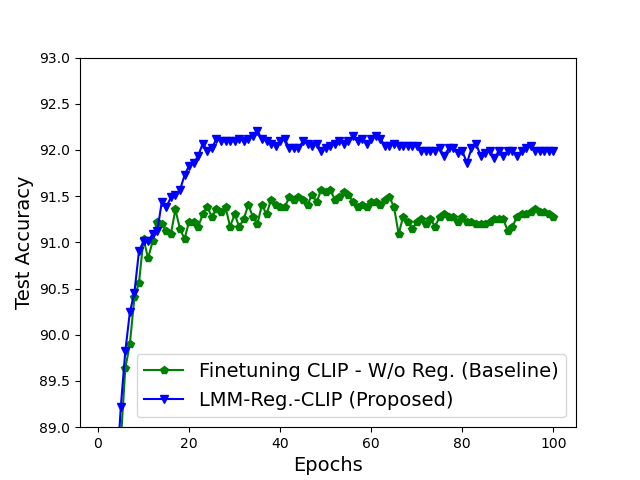}
\caption{UCF-101 dataset.}
\label{fig:ucf}
    \end{subfigure}
    \hfill
    \begin{subfigure}{0.3\textwidth}
\centering
\includegraphics[width=\textwidth]{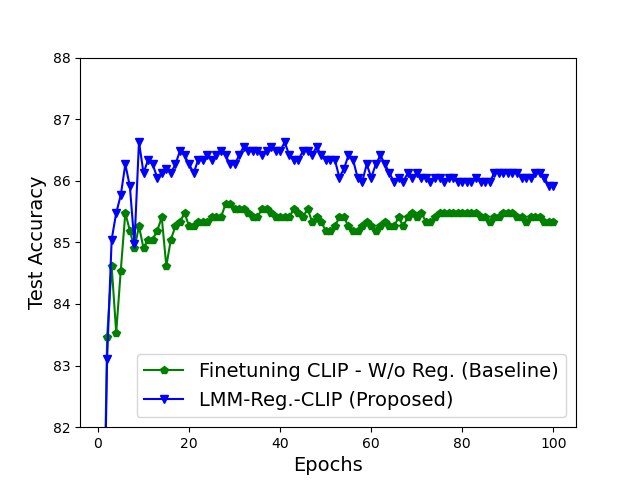}
\caption{ERA dataset.}
\label{fig:era}
    \end{subfigure}
    \hfill
    \begin{subfigure}{0.3\textwidth}
\centering
\includegraphics[width=\textwidth]{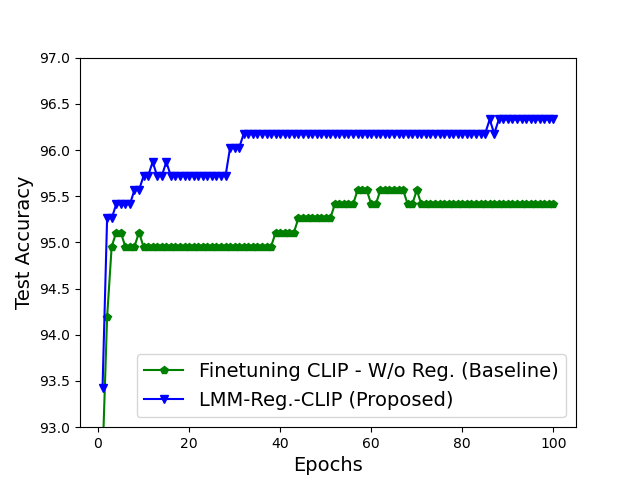}
\caption{BAR dataset}
\label{fig:bar}
\end{subfigure}
      \caption{Test accuracy throughout the training epochs for the proposed method against baseline.}
    \label{fig:plots}  
\end{figure*}

The experimental results of the proposed regularization method for the three considered datasets are provided in Table \ref{tab:1}. We compare the proposed method, abbreviated as LMM-Reg.-CLIP, with the baseline training which consists in finetuning the CLIP's image encoder without applying the proposed regularization objective, using the same experimental setup. Best results are printed in bold. As it can be observed, LMM-Reg.-CLIP considerably improves the baseline performance. Correspondingly, in Fig. \ref{fig:plots} we illustrate the test accuracy throughout the training epochs for the three considered datasets, where it can be observed that the proposed method provides steadily advanced performance as compared to the baseline training. 

\begin{table}[!ht]
\begin{center}
\caption{Test accuracy (\%) of the proposed method against baseline.} \label{tab:1}
\begin{tabular}{|c|c|c|c|}
  \hline
  \bf{Method} & \bf{UCF-101} & \bf{ERA} & \bf{BAR}
  \\ 
  \hline
Finetuning w/o Reg. (Baseline)  &  91.567 & 85.621 &  95.565\\ \hline
LMM-Reg.-CLIP (Proposed)  &  \bf{92.202} & \bf{86.628}&  \bf{96.330}\\ \hline
\end{tabular}
\end{center}
\end{table}

Subsequently, in Table \ref{tab:comparisons} we present the comparisons of the proposed LMM-Reg.-CLIP method against CLIP-based state-of-the-art methods. Specifically, apart from the most straightforward comparison against the finetuning with the class labels without regularization, presented in Table \ref{tab:1}, we also compare the performance with CLIP utilizing only the image embeddings \cite{radford2021learning}, while another straightforward comparison is against \cite{tzelepi2024exploiting}, where both image and LMM-generated text embeddings are used, concatenated, for feeding a linear classifier. Furthermore, we include other relevant works that use CLIP and LLMs/LMMs. All the models include training. As demonstrated, the proposed method provides state-of-the-art performance.

\begin{table}
\centering
\caption{Comparisons with CLIP-based approaches in terms of test accuracy (\%).} \label{tab:comparisons}
  \begin{tabular}{|c|c|c|c|}
  \hline
  \bf{Method} & \bf{UCF-101} & \bf{ERA} & \bf{BAR} \\ 
  \hline
  CLIP-A-self \cite{maniparambil2023enhancing} & 84.1 & - & - \\ \hline
  CoOp \cite{zhou2022learning} & 84.69 & - & - \\ \hline
  CoCoOp \cite{zhou2022conditional} & 82.33 & - & - \\ \hline
  LaBo \cite{yang2023language} & 90.67 & - & - \\ \hline
  CLIP (image emb.) \cite{radford2021learning} & 89.981 & 84.472 & 94.801 \\ \hline
  CLIP (image \& text emb.) \cite{tzelepi2024exploiting} & 91.753 & 85.909 & 95.566 \\ \hline
  LMM-Reg.-CLIP (Proposed) & \bf{92.202} & \bf{86.628}&  \bf{96.330}\\  \hline
  \end{tabular}
\end{table}

%
%

\section{Conclusions}\label{sec:con}
In this paper we addressed image classification tasks using an LMM-based regularization method. The proposed method uses first an LMM to extract semantic descriptions for the images of the dataset. It uses, then, the CLIP's text encoder to extract the corresponding text embeddings and compute mean class descriptions. Then, instead of training the image encoder of CLIP with class labels using a supervised loss, it introduces an additional regularization objective that forces the embeddings at the output of the image encoder to become similar to the LMM-generated mean class descriptions. In this way, the discriminability of the aforementioned embeddings is enhanced, leading to advanced classification performance, as it was experimentally validated on the three datasets.  

\section*{Acknowledgments}
This work has been funded by the European Union as part of
the Horizon Europe Framework Program, under grant agreement 101070109 (TransMIXR).

\begin{thebibliography}{10}

\bibitem{radford2021learning}
Alec Radford, Jong~Wook Kim, Chris Hallacy, Aditya Ramesh, Gabriel Goh, Sandhini Agarwal, Girish Sastry, Amanda Askell, Pamela Mishkin, Jack Clark, et~al.
\newblock Learning transferable visual models from natural language supervision.
\newblock In {\em International conference on machine learning}, pages 8748--8763. PMLR, 2021.

\bibitem{ao2023gesturediffuclip}
Tenglong Ao, Zeyi Zhang, and Libin Liu.
\newblock Gesturediffuclip: Gesture diffusion model with clip latents.
\newblock {\em arXiv preprint arXiv:2303.14613}, 2023.

\bibitem{zanella2024delving}
Luca Zanella, Benedetta Liberatori, Willi Menapace, Fabio Poiesi, Yiming Wang, and Elisa Ricci.
\newblock Delving into clip latent space for video anomaly recognition.
\newblock {\em Computer Vision and Image Understanding}, page 104163, 2024.

\bibitem{zhao2023survey}
Wayne~Xin Zhao, Kun Zhou, Junyi Li, Tianyi Tang, Xiaolei Wang, Yupeng Hou, Yingqian Min, Beichen Zhang, Junjie Zhang, Zican Dong, et~al.
\newblock A survey of large language models.
\newblock {\em arXiv preprint arXiv:2303.18223}, 2023.

\bibitem{yin2023survey}
Shukang Yin, Chaoyou Fu, Sirui Zhao, Ke~Li, Xing Sun, Tong Xu, and Enhong Chen.
\newblock A survey on multimodal large language models.
\newblock {\em arXiv preprint arXiv:2306.13549}, 2023.

\bibitem{zhu2023minigpt}
Deyao Zhu, Jun Chen, Xiaoqian Shen, Xiang Li, and Mohamed Elhoseiny.
\newblock Minigpt-4: Enhancing vision-language understanding with advanced large language models.
\newblock {\em arXiv preprint arXiv:2304.10592}, 2023.

\bibitem{tzelepi2019graph}
Maria Tzelepi and Anastasios Tefas.
\newblock Graph embedded convolutional neural networks in human crowd detection for drone flight safety.
\newblock {\em IEEE Transactions on Emerging Topics in Computational Intelligence}, 5(2):191--204, 2019.

\bibitem{zhou2022learning}
Kaiyang Zhou, Jingkang Yang, Chen~Change Loy, and Ziwei Liu.
\newblock Learning to prompt for vision-language models.
\newblock {\em International Journal of Computer Vision}, 130(9):2337--2348, 2022.

\bibitem{zhou2022conditional}
Kaiyang Zhou, Jingkang Yang, Chen~Change Loy, and Ziwei Liu.
\newblock Conditional prompt learning for vision-language models.
\newblock In {\em Proceedings of the IEEE/CVF Conference on Computer Vision and Pattern Recognition}, pages 16816--16825, 2022.

\bibitem{yang2023language}
Yue Yang, Artemis Panagopoulou, Shenghao Zhou, Daniel Jin, Chris Callison-Burch, and Mark Yatskar.
\newblock Language in a bottle: Language model guided concept bottlenecks for interpretable image classification.
\newblock In {\em Proceedings of the IEEE/CVF Conference on Computer Vision and Pattern Recognition}, pages 19187--19197, 2023.

\bibitem{brown2020language}
Tom Brown, Benjamin Mann, Nick Ryder, Melanie Subbiah, Jared~D Kaplan, Prafulla Dhariwal, Arvind Neelakantan, Pranav Shyam, Girish Sastry, Amanda Askell, et~al.
\newblock Language models are few-shot learners.
\newblock {\em Advances in neural information processing systems}, 33:1877--1901, 2020.

\bibitem{maniparambil2023enhancing}
Mayug Maniparambil, Chris Vorster, Derek Molloy, Noel Murphy, Kevin McGuinness, and Noel~E O'Connor.
\newblock Enhancing clip with gpt-4: Harnessing visual descriptions as prompts.
\newblock In {\em Proceedings of the IEEE/CVF International Conference on Computer Vision}, pages 262--271, 2023.

\bibitem{achiam2023gpt}
Josh Achiam, Steven Adler, Sandhini Agarwal, Lama Ahmad, Ilge Akkaya, Florencia~Leoni Aleman, Diogo Almeida, Janko Altenschmidt, Sam Altman, Shyamal Anadkat, et~al.
\newblock Gpt-4 technical report.
\newblock {\em arXiv preprint arXiv:2303.08774}, 2023.

\bibitem{tzelepi2024exploiting}
Maria Tzelepi and Vasileios Mezaris.
\newblock Exploiting lmm-based knowledge for image classification tasks.
\newblock In {\em International Conference on Engineering Applications of Neural Networks}, pages 166--177. Springer, 2024.

\bibitem{tzelepi2024disturbing}
Maria Tzelepi and Vasileios Mezaris.
\newblock Disturbing image detection using lmm-elicited emotion embeddings.
\newblock In {\em Proceedings of LVLM Workshop @ 2024 IEEE Int. Conf. on Image Processing (ICIP 2024), Abu Dhabi, UAE, Oct. 2024.}

\bibitem{soomro2012ucf101}
Khurram Soomro, Amir~Roshan Zamir, and Mubarak Shah.
\newblock Ucf101: A dataset of 101 human actions classes from videos in the wild.
\newblock {\em arXiv preprint arXiv:1212.0402}, 2012.

\bibitem{eradataset}
L.~Mou, Y.~Hua, P.~Jin, and X.~X. Zhu.
\newblock {ERA: A dataset and deep learning benchmark for event recognition in aerial videos}.
\newblock {\em IEEE Geoscience and Remote Sensing Magazine}, in press.

\bibitem{nam2020learning}
Junhyun Nam, Hyuntak Cha, Sungsoo Ahn, Jaeho Lee, and Jinwoo Shin.
\newblock Learning from failure: De-biasing classifier from biased classifier.
\newblock {\em Advances in Neural Information Processing Systems}, 33:20673--20684, 2020.

\end{thebibliography}

\end{document}